\documentclass[conference,twocolumn,10pt]{IEEEtran}

\usepackage{setspace}
\usepackage{amsmath,amsthm}
\usepackage{amssymb}
\usepackage{enumerate}
\usepackage{bm} 
\usepackage{graphicx}
\usepackage{subcaption}
\usepackage{color}
\graphicspath{{Figures/}}
\usepackage{lipsum}
\usepackage{stfloats}
\usepackage{multicol}
\usepackage{multirow}
\usepackage{cite}
\allowdisplaybreaks

\newcommand{\unif}{\text{unif}}
\newcommand{\eg}{\emph{e.g.,}}

\newtheorem{lemma}{Lemma}
\newtheorem{theorem}{Theorem}
\newtheorem{definition}{Definition}

\newtheorem{remark}{Remarks}

\newcommand{\vol}{\text{vol}}
\newcommand{\diam}{\text{diam}}
\newcommand{\real}{\mathbb{R}}
\newcommand{\argmin}{\text{argmin}}

\newcommand{\bk}{\mathbf{k}}
\newcommand{\cE}{\mathcal{E}}

\newcommand{\fmarg}{\phi_{\text{marg}}}
\newcommand{\cN}{\mathcal{N}}
\newcommand{\relu}{\text{ReLU}}
\newcommand{\pmarg}{P_{\text{marg}}}
\newcommand{\sfm}{\mathsf{M}}

\title{Approximating Probability Distributions by ReLU Networks}
\author{
\IEEEauthorblockN{Manuj Mukherjee$^\dag$} \hspace{1cm} \and \IEEEauthorblockN{Aslan Tchamkerten$^\ddag$} \hspace {1cm} \and \IEEEauthorblockN{Mansoor Yousefi$^\ddag$}
}

\begin{document}
\maketitle

\renewcommand{\thefootnote}{}
\footnotetext{
\noindent $^\dag$M.\ Mukherjee is with The Alexander Kofkin Faculty of Engineering, Bar Ilan University, Ramat Gan, Israel. Email: mukherm@biu.ac.il.

\smallskip

$^\ddag$A.\ Tchamkerten and M.\ Yousefi are with Telecom Paris, Institut Polytechnique de Paris, Palaiseau, France. Email: \{aslan.tchamkerten, yousefi\}@telecom-paris.fr.
}

\renewcommand{\thefootnote}{\arabic{footnote}}

\begin{abstract}
How many neurons are needed to approximate a target probability distribution using a neural network with a given input distribution and approximation error? This paper examines this question for the case when the input distribution is uniform, and the target distribution belongs to the class of histogram distributions. We obtain a new upper bound on the number of required neurons, which is strictly better than previously existing upper bounds. The key ingredient in this improvement is an efficient construction of the neural nets representing piecewise linear functions. We also obtain a lower bound on the minimum number of neurons needed to approximate the histogram distributions. 
\end{abstract}

\section{Introduction}\label{sec:intro}

Neural networks have long been used as classifiers, whose underlying task is function computation. The theory of function approximation via neural networks is now well-studied \cite{Barron, Tel16, Yarot, MAM, ABMM, EPGB, Arora}. Recently, through the development of \emph{variational autoencoders} \cite{KW} and \emph{generative adversarial networks (GAN)} \cite{IG}, neural networks are also being used to approximate probability distributions.

In the distribution approximation setting, a neural network is provided with an input random variable $X$ (seed) distributed according to some distribution that can be easily generated, \eg\ Gaussian or uniform. The task of the neural network is to compute a function $f$ such that the distribution of $f(X)$ closely approximates a target distribution. 

A natural question to ask is the following. Given a seed distribution and a target distribution, what is the minimum number of neurons, \textit{i.e.}, the minimum size of the neural network, needed to approximate the target distribution within a given accuracy? The  work of Bailey and Telgarsky \cite{BT} tackled this question for a variety of seed-target pairs. In particular, given a network of a fixed depth, \cite{BT} provided an upper bound on the minimum network size needed to approximate the target $\unif[0,1]^h$ using the seed $\unif[0,1]^l$, where $h>l$, $h,l\in\mathbb N$. The authors also showed that this upper bound is tight upto constant factors. 

Building upon \cite{BT}, \cite{PSB} examined the task of approximating generalisations of the uniform distribution, referred to as \emph{histogram distributions} (see Definition~\ref{def:hist}). The authors in \cite{PSB} constructed neural networks which are able to approximate $n$-tiled histogram distributions using uniform seeds upto a given accuracy. However, \cite{PSB} does not comment on whether their construction is optimal with respect to the size of the network.

In this work, we examine the optimality of the construction in \cite{PSB}. Given uniform seed, approximation error $\epsilon$, and a target histogram distribution $P$ of tile parameter $n$ (see Definition~\ref{def:hist}), we construct a  neural net which approximates $P$ within $\epsilon$. Compared to the construction in \cite{PSB} (see Table~\ref{tab:compachiev}),  our construction uses strictly fewer neurons in various regimes of $\epsilon$ and $n$. For example, in the regime where $\epsilon$ is fixed and $n\to\infty$, we achieve a network size of $O(n^{\frac{3}{2}})$ improving upon $\Theta(n^2)$ in \cite{PSB}. In other regimes, our construction uses at most the same number of neurons as that of \cite{PSB}. The key ingredient that we used to reduce the number of neurons is an efficient construction of the neural networks that compute piecewise affine functions studied in \cite{DDFHP}. Instead of representing a piecewise affine function by a straightforward neural net of depth 2 and size linear in the number of affine pieces, \cite{DDFHP} constructs a deeper neural network computing the same function. Suitably choosing the depth of this neural net, we are able to reduce the size of the network compared to the straightforward network. 

To examine the optimality of our construction, we developed a lower bound on the number of neurons needed by neural networks of a fixed depth to approximate histogram distributions. The bound is developed by extending the techniques of \cite{BT}. In certain regimes, such as when $\epsilon\to 0$ and $n$ is fixed, both our construction and that of \cite{PSB} turn out to be optimal. However, in most other regimes, the lower bound on the number of neurons does not match the upper bound in our construction. The optimal construction remains unknown.

The rest of the paper is organised as follows. The problem of approximating distributions and our main result are formally stated in Section~\ref{sec:prelims}, along with the comparison of our construction with that of \cite{PSB}. Section~\ref{sec:proof} proves the main result by constructing a neural network of reduced size. Section~\ref{sec:conv} states the lower bound on the minimum size needed to approximate histogram distributions. The paper is concluded in Section~\ref{sec:conc}.

\section{Preliminaries}\label{sec:prelims}

A \emph{fully-connected feedforward rectified linear unit (ReLU)} neural network is formally defined as follows.
\begin{definition}\label{def:nn}
A map $\phi:\mathbb{R}^{m_0}\to\real^{m_L}$ is a fully connected feedforward ReLU network of depth $L$ and size $N$, if there exists integers $m_1,m_2,\ldots,m_{L-1}>0$, such that 
\begin{itemize}
\item $\phi(x)=\mathsf{M}_L(\relu(\mathsf{M}_{L-1}(\relu(\ldots\relu(\mathsf{M}_1(x))$, where $\mathsf{M}_i:\real^{m_{i-1}}\to\real^{m_i}, 1\leq i\leq L$, are affine maps, and $\relu(x)=(\max\{0,x_i\}:1\leq i\leq m)$, for any $x\in\real^m$.
\item $\sum_{i=1}^L m_i=N$. In other words, the size is simply the total number of neurons.
\end{itemize}

\end{definition}
Notice that widths need not be the same.
Throughout this paper, we consider neural networks 
with input dimension $m_0=1$ and output dimension $m_L=2$,\footnote{We restrict attention to two-dimensional output for simplicity, since our goal is to show how a specific construction of neural nets representing linear splines (see Lemma~\ref{lem:spline}) can be used to reduce the overall size of the network. The results can be extended to higher dimensions, \textit{i.e.}, $m_L>2$, however, the proofs (especially of Lemma~\ref{lem:map}) become cumbersome.}  unless otherwise stated. Denote the set of all neural networks with size $N$ and depth $L$ by $\cN(N,L)$.


The goal of this work is to approximate \emph{$n$-tiled histogram} probability distributions at the output. These are  distributions that are obtained by quantizing $[0,1]^2$ into $n^2$ uniform squares and assigning probability masses to each of them: 
\begin{definition}
\label{def:hist}
A probability distribution supported on $[0,1]^2$ is an $n$-tiled histogram if its pdf $P$ is
\begin{IEEEeqnarray}{rCl}
P(x)=\sum_{\bk \in K} w_\bk\mathbf{1}\{x\in c_\bk\},
\end{IEEEeqnarray}
where $K=\bigl\{ (k_1,k_2): k_i=0,\cdots,n-1\bigr\}$, $c_\bk=[\frac{k_1}{n},\frac{k_1+1}{n}] \times [\frac{k_2}{n},\frac{k_2+1}{n}]$, $w_{\bk}>0$, and $\sum_{\bk}w_\bk=n^2$. The set of all $n$-tiled histogram distributions supported on $[0,1]^2$ is denoted by $\cE[0,1]_n^2$. 
\end{definition}

As a distance between distributions we choose the \emph{Wasserstein distance}.
\begin{definition}\label{def:wasser}
For a pair of distributions $P$ and $Q$ supported on $\real^2$, denote by $\Pi(P,Q)$ the set of \emph{couplings}, \textit{i.e.}, joint distributions on $\real^2\times\real^2$ whose marginals are respectively $P$ and $Q$. Then, the Wasserstein distance between $P$ and $Q$ is given by 
$$
\mathsf{W}(P,Q)\triangleq\inf_{\pi\in\Pi(P,S)}\int_{\real^2\times\real^2}||x-y||d\pi(x,y),
$$
where $||\cdot||$ denotes the $\ell_2$-norm.
\end{definition}

We shall denote the uniform distribution on $[0,1]$ by $U$. Given any random variable $X\sim P(x)$ and any function $f$, the distribution of $Y=f(X)$ shall be denoted by $f\# P$. We say that a neural network $\phi:\real\to\real^2$ $\epsilon$-approximates $P\in\cE[0,1]_n^2$ with uniform seed if $\mathsf{W}(P,\phi\# U)\leq\epsilon$.

\begin{remark}
The choice of Wasserstein distance over more conventional measures like total variation (TV) is dictated by the fact that any neural network $\phi:\real\to\real^2$ and any $P\in\cE[0,1]_n^2$ satisfies $\text{TV}(P,\phi\# U)=1$. This is due to the fact that $\phi(U)$ is always supported on a collection of line segments in $\real^2$ where $P$ puts no mass. Refer to \cite[Section~2]{WGAN} for a detailed discussion. 
\end{remark}
\begin{definition}
\label{def:achiev}
We say that $N$ is an achievable size for $\epsilon$-approximating $\cE[0,1]_n^2$ with uniform seed if, for any $P\in\cE[0,1]_n^2$, there exists an integer $L>0$ and a neural net $\phi\in\cN(N,L)$ which $\epsilon$-approximates $P$.

Denote by $N^*(n,\epsilon)$ the minimum network size needed for $\epsilon$-approximating $\cE[0,1]_n^2$ with uniform seed.
\end{definition}

Our main result, stated next, gives an upper bound on $N^*(n,\epsilon)$. 


\begin{theorem}
\label{th:main}
Fix $\epsilon>0$. The minimum number of neurons needed to $\epsilon$-approximate $\cE[0,1]_n^2$ satisfies
$$
N^*(n,\epsilon)\leq O(n^{\frac{3}{2}})+\max\biggl\{1,\biggl\lceil\log\biggl(\frac{2\sqrt{2}}{n\epsilon}\biggr)\biggr\rceil\biggr\}O(n).
$$
\end{theorem}
We prove Theorem~\ref{th:main} in  Section~\ref{sec:proof} where we 
\begin{table*}
\centering
\begin{tabular}{c|c|c}
Regime & Our construction & Construction in \cite{PSB}\\
\hline
\hline
$n$ fixed & $N=\Theta(\log(\frac{1}{\epsilon}))$ & $N=\Theta(\log(\frac{1}{\epsilon}))$\\
$\epsilon\to0$ & \fontsize{8}{8}{$L=\Theta(\log(\frac{1}{\epsilon}))$} & \fontsize{8}{8}{$L=\Theta(\log(\frac{1}{\epsilon}))$}\\
\hline
$n\to\infty$ & $N=\Theta(n^{\frac{3}{2}})$ & $N=\Theta(n^2)$\\
$\epsilon$ fixed & \fontsize{8}{8}{$L=\Theta(n)$} & \fontsize{8}{8}{$L=\Theta(1)$}\\
\hline 
$n\to\infty$ & $N=\Theta(n^{\frac{3}{2}})$ & $N=\Theta(n^2)$\\
$\epsilon=\Theta(n^{-\theta}), \theta>1$ & \fontsize{8}{8}{$L=\Theta(n\log n)$} & \fontsize{8}{8}{$L=\Theta(\log n)$}\\
\hline
$n\to\infty$ & $N=\Theta(n^2)$ & $N=\Theta(n^2)$\\
$\epsilon=\Theta(2^{-n})$ & \fontsize{8}{8}{$L=\Theta(n^2)$} & \fontsize{8}{8}{$L=\Theta(n)$}\\
\end{tabular}
\caption{Table comparing the sizes and depths of neural networks constructed by us and \cite{PSB} for different regimes of $n$ and $\epsilon$. Our construction uses deeper nets to cut down on the size of the network.}
\label{tab:compachiev}
\end{table*}
construct a neural net of size $N=\Theta(n^{\frac{3}{2}})+\max\biggl\{1,\biggl\lceil\log\biggl(\frac{2\sqrt{2}}{n\epsilon}\biggr)\biggr\rceil\biggr\}\Theta(n)$ and depth $L=\Theta(n)+n\max\biggl\{1,\biggl\lceil\log\biggl(\frac{2\sqrt{2}}{n\epsilon}\biggr)\biggr\rceil\biggr\}$, which $\epsilon$-approximates distributions in $\cE[0,1]_n^2$.

In Table~\ref{tab:compachiev} we compare our construction to that of \cite{PSB}.\footnote{The result in Theorem~4.4 of \cite{PSB} is originally stated in terms of depth and a parameter called `connectivity'. The bound on the size $N^*(n,\epsilon)$ stated here can be obtained by looking at the construction used in the proof.}Our construction performs at least as good as the construction of \cite{PSB} in all regimes of $\epsilon$ and $n$, and sometimes better.

\section{The achievability Part: Proof of Theorem~\ref{th:main}}\label{sec:proof}

The proof is based on two steps.
Given $P\in\cE[0,1]_n^2$ we first construct a piece-wise linear function $\phi:\real\to\real^2$ which satisfies $\mathsf{W}(P,\phi\# U)\leq\epsilon$. This step is based on \cite[Theorem~4.4]{PSB}. In the second step, we construct a ReLU neural network with desired size $N$ and that realizes $\phi$ exactly.
This  construction is based on results from \cite{DDFHP} for representing \emph{free-knot linear splines} and sums of functions using neural networks. The improvement over \cite{PSB} results from this second step.

\subsection{Constructing a piece-wise linear function $\phi$}\label{sec:pushforward}

We will keep this section short as it covers material from \cite{PSB}. We shall state only the relevant details that will be required in order to state the function $\phi$.

Define the triangular function $g:[0,1]\to[0,1]$ 
\begin{equation}
g(x)=
\begin{cases}
2x, & x\in[0,\frac{1}{2}]\\
2-2x, & x\in[\frac{1}{2},1]. \label{eq:g}
\end{cases}
\end{equation}
Denote the composition of $s$ triangular functions by $g_s\triangleq\underbrace{g_1\circ g_1\circ\ldots\circ g_1}_{s \text{ times}}$. The resulting function $g_s$ is a ``sawtooth" function 
\begin{equation}
g_s(x)=
\begin{cases}
2^s(x-i2^{-s+1}),  &\\
\hspace{0.5cm}   x\in [2i ,2i+1]\times 2^{-s}, \quad\forall i\in I_s,  &\\
-2^s(x-(i+1)2^{-s+1}), & \\
\hspace{0.5cm}x\in[2i+1,2i+2]\times 2^{-s}, \quad \forall i\in I_s,&
\end{cases}\label{eq:saw}
\end{equation}
where $I_s= \{0,1,\ldots, 2^{s-1}-1\}$.

Denote by $\Sigma_m$ the set of \emph{continuous piecewise affine functions} $h:[0,1]\to[0,1]$ having $m$ breakpoints in $(0,1)$. 
Constructing $\phi$ when $m_L=1$ is straightforward.
\begin{lemma}\label{lem:push1d}
For every $P\in\cE[0,1]_n$, there exists an $f\in\Sigma_{n-1}$ satisfying $f\# U=P$.
\end{lemma} 
\begin{IEEEproof}
The result follows by noting that $F_P^{-1}\in\Sigma_{n-1}$ if $P\in\cE[0,1]_n$, where $F_P$ denotes the cumulative distribution function of $P$. Since $F_P^{-1}\# U=P$, $f=F^{-1}_P$. 
\end{IEEEproof}

To generalize the construction from $m_L=1$ to $m_L=2$, consider
$p(y_1,y_2)= p(y_1) p(y_2|y_1)$. The component $p(y_1)$ can be constructed from Lemma~\ref{lem:push1d} by marginalization.
Let the weights of $P$ for each part $c_{k_1,k_2}=[\frac{k_1}{n},\frac{k_1+1}{n}]\times[\frac{k_2}{n},\frac{k_2+1}{n}]$ be denoted by $w_{k_1,k_2}^P$. Denote by $\pmarg$ the $n$-tiled histogram distribution on $[0,1]$ whose weight are given by $w_k^{\pmarg}=\sum_{j=0}^{n-1}w_{k,j}^P$, for all $k\in\{0,1,\ldots,n-1\}$. In other words, $\pmarg$ is the marginal distribution of $P$ along the first coordinate. Let $\fmarg\in\Sigma_{n-1}$ be the function given by Lemma~\ref{lem:push1d} which satisfies $\fmarg\# U=\pmarg$.
For the component $p(y_2|y_1)$, let $P_i$, $0\leq i\leq n-1$, be the $n$-tiled histogram distributions in $[0,1]$ with weights given by $w_k^{P_i}=\frac{nw_{i,k}^P}{\sum_{j=0}^{n-1}w_{j,k}^P}$, for all $k\in\{0,1,\ldots,n-1\}$. The distribution $P_i$ are simply the marginals of $P$ along the second coordinate conditioned on the fact that the first coordinate lies in $[\frac{i}{n},\frac{i+1}{n}]$. Let $\phi_i\in\Sigma_{n-1}$ be the functions given by Lemma~\ref{lem:push1d}, which satisfy $\phi_i\# U=P_i$, for all $0\leq i\leq n-1$. 

With these two ingredients, we can now present the function $\phi:[0,1]\to[0,1]^2$ that approximates an $n$-tiled histogram distribution $P$.
\begin{lemma}\label{lem:map}
Let $P\in\cE[0,1]^2_n$. The corresponding function $\phi:[0,1]\to[0,1]^2$ defined by
\begin{IEEEeqnarray}{rCl}
\phi(x)=(\fmarg(x),\sum_{i=0}^{n-1}\phi_i(g_s(n\fmarg(x)-i))),
\label{eq:phi(x)}
\end{IEEEeqnarray}
where $g_s(.)$ is defined in \eqref{eq:saw}, satisfies $\mathsf{W}(P,\phi\# U)\leq\frac{2\sqrt{2}}{n2^s}$ for any $s\in\mathbb N$.
\end{lemma}
 
\begin{IEEEproof}
See the proof of Theorem~4.4 of \cite{PSB}.
\end{IEEEproof}

Hence, by choosing $s=\max\{1,\lceil\log(\frac{2\sqrt{2}}{n\epsilon})\rceil\}$, we have $\mathsf{W}(\phi\# U,P)\leq \epsilon$ for any $P\in\cE[0,1]_n^2$. 


\subsection{Constructing a ReLU neural network}\label{sec:const}

In this section we construct a ReLU neural net that realises the function $\phi$ given by Lemma~\ref{lem:map}.
The proof is based on realizing each component in \eqref{eq:phi(x)} by a ReLU network efficiently.
We begin by proving the required lemmas.  

\begin{lemma}[Pass Lemma]\label{lem:pass}
If $f:[0,1]\to[0,1]$ satisfies $f\in\cN(N,L)$, then for any $L'\geq L$, $f\in\cN(N+L'-L,L')$.
\end{lemma}
\begin{IEEEproof}
This is possible by simply adding $L'-L$ layers each of which contain only one neuron, and define the maps $\mathsf{M}_i$, for all $L+1\leq i\leq L'$ to be identity maps. Note that the ReLU of the added layers have no effect since the function takes values in $[0,1]$.
\end{IEEEproof}

\begin{lemma}[Parallel Lemma]\label{lem:parallel}
Let $f,g:[0,1]\to[0,1]$ satisfy $f\in\cN(N_1,L_1)$ and $g\in\cN(N_2,L_2)$. Then the function $h:[0,1]\to[0,1]^2$ given by $h=(f,g)$ satisfies $h\in\cN(N_1+N_2+|L_1-L_2|,\max\{L_1,L_2\})$.
\end{lemma}

\begin{IEEEproof}
Without loss of generality, assume $L_1\geq L_2$. Then, by the Lemma~\ref{lem:pass}, we have $g\in\cN(N_2+|L_1-L_2|,L_1)$. To get the network $h$, we simply stack the net $f\in\cN(N_1,L_1)$ on top of net $g\in\cN(N_2+|L_1-L_2|,L_1)$, and ensure that the weights of the connections between neurons in $f$ to neurons in $g$ is zero. Formally, let the affine maps for $f$ and $g$ be denoted by $\mathsf{M}^{(f)}_i$ and $\mathsf{M}^{(g)}_i$, for all $1\leq i\leq L_1$. Then, define the new neural net $h:[0,1]\to[0,1]^2$ by using the concatenated affine maps $\mathsf{M}_i=(\mathsf{M}^{(f)}_i,\mathsf{M}^{(g)}_i)$, for all $1\leq i\leq L_1$. It is easy to see that $h=(f,g)$ and $h\in\cN(N_1+N_2+|L_1-L_2|,\max\{L_1,L_2\})$.
\end{IEEEproof}

\begin{lemma}[Compose Lemma]\label{lem:compose}
Let $f,g:\mathbb{R}\to\mathbb{R}$. If $f\in\cN(N_1,L_1)$ and $g\in\cN(N_2,L_2)$, then for any $p,q\in\real$, $f\circ (pg+q)\in\cN(N_1+N_2-1,L_1+L_2-1)$.
\end{lemma}

\begin{IEEEproof}
We shall use superscripts $^{(f)}$ and $^{(g)}$ to denote affine maps from the neural nets realising $f$ and $g$. The network $f\circ (pg+q)$ is simply generated by adding the net $f$ in series with net $g$, deleting the output node of $g$, and connecting the neurons from layer $L_2-1$ of $g$ to the first layer of $f$ using suitable weights. 

Formally, let the affine maps $\sfm_{L_2}^{(g)}$ and $\sfm_1^{(f)}$ be expressed as $\sfm_{L_2}^{(g)}(x)=A^{(g)}x+b^{(g)}$ and $\sfm_1^{(f)}(x)=A^{(f)}x+b^{(f)}$, where $A^{(g)}\in\real^{1\times m_{L_2-1}^{(g)}}$, $A^{(f)}\in\real^{m_1^{(f)}\times1}$, $b^{(g)}\in\real$, $b^{(f)}\in\real^{m_1^{(f)}}$. Now, define the new affine map $\sfm:\real^{m_{L_2-1}^{(g)}}\to\real^{m_1^{(f)}}$ as $\sfm(x)=Ax+b$, where $A\in\real^{m_1^{(f)}\times m_{L_2-1}^{(g)}}$, $b\in\real^{m_1^{(f)}}$, and $A=pA^{(f)}A^{(g)}$, $b=(pb^{(g)}+q)A^{(f)}+b^{(f)}$. Now, consider the net of depth $L_1+L_2-1$ defined by the affine maps $\sfm_i=\sfm_i^{(g)}$, for all $1\leq i\leq L_2-1$, $\sfm_{L_2}=\sfm$, and $\sfm_{L_2+i}=\sfm^{(f)}_{1+i}$, for all $1\leq i\leq L_1-1$. It is easy to see that this net has size $N_1+N_2-1$ and realises the map $f\circ(pg+q)$.
\end{IEEEproof}

Next, we state a couple of important technical result from \cite{DDFHP} dealing with representing functions from $\Sigma_m$ and sums of functions.\footnote{The results in \cite{DDFHP} are stated in terms of width of network given by $W=\max_{1\leq i\leq L_1}m_i$. Moreover, by the convention of \cite{DDFHP}, the depth is $L-1$ instead of $L$. We have therefore suitably modified the statements according to our convention, and also expressed the statements in terms of size and not width.} 

\begin{lemma}[Spline Lemma]\label{lem:spline}
Let $f\in\Sigma_m$. Then $f$ can be realised by a ReLU network with $L=\max\biggl\{2,2\biggl\lceil\frac{m}{(W-2)\lfloor\frac{W-2}{6}\rfloor}\biggr\rceil+1\biggr\}$ and $N=W(L-1)+1$, for any $W\geq 8$.
\end{lemma}

\begin{IEEEproof}
See Theorem~3.1 of \cite{DDFHP}.
\end{IEEEproof}

Note that  $f\in\Sigma_m$ can always be realised using a depth-2 ReLU network of size $m+3$ in a straightforward manner. The spline lemma, however, uses a deeper net and realizes $f\in\Sigma_m$ more efficiently, thereby saving up on the number of neurons. 
Below we explain how the Spline Lemma does this.

The Spline Lemma is proved by viewing the set of piecewise affine functions with $m$ breakpoints  as a vector space of dimension $m$, and constructing a suitable basis for this space. This vector space representation is more efficient than direct computation of the  sum \eqref{eq:phi(x)}. As a result, 
any piecewise affine function $h$ of $m_W\triangleq(W-2)\lfloor\frac{W-2}{6}\rfloor$ breakpoints, where $W\geq8$ is a parameter, can be expressed as a sum $h(x)=\sum_{k=1}^{W-2}T_k$, where $T_k=\sum_{j=1}^{W-2}c_{j,k}\relu(x-\beta_j)$, where $c_{j,k},\beta{j}\in\real$.
The function $h=\sum_{k=1}^{W-2}T_k$ can be realised by a network of three layers with the two intermediate layers having $W-2$ nodes each, and given that the second intermediate layer is activation free. A simple argument (see \cite[Remark~3.1]{DDFHP}) shows that this artificial technical requirement can be removed by simply adding two more neurons to both the intermediate layers, and adjusting some weights suitably. Now, any $f\in\Sigma_m$ can be expressed as $f=\sum_{i=1}^{\lceil\frac{m}{m_W}\rceil}h_i$, where $h_i$ are piecewise affine functions with $m_W$ breakpoints, and can be realised using the neural networks of size $2(W-2)+1$, depth $3$, and activation free second layer as described above. The result in the Spline Lemma then follows by realising the sum in $f$ using a network constructed according to the following Add Lemma (Lemma~\ref{lem:seradd}). The proof of the Spline Lemma includes the technical argument to remove the artificial restriction of the second layer of each $h_i$ being activation free.    

\begin{lemma}[Add Lemma]\label{lem:seradd}
Consider $\ell$ functions $f_i:[0,1]\to\mathbb{R}$, $1\leq i\leq\ell$, with $f_i\in(N_i,L_i)$. Then the sum $f=\sum_{i=1}^\ell f_i$ satisfies $f\in\cN(\sum_{i=1}^\ell(N_i+2L_i-2)-\ell+1,\sum_{i=1}^\ell L_i-\ell+1)$.  
\end{lemma}

\begin{IEEEproof}
The proof involves putting the networks in series and adding a pair of nodes to each intermediate layer, one to pass the input, and the other to pass the partial sum. See Proposition~4.2(ii) and Remark~3.1 of \cite{DDFHP} for details.
\end{IEEEproof}

With these lemmas in hand, we begin the construction of a net for $\phi$. Firstly, recall that $\fmarg, \phi_i\in\Sigma_{n-1}$, for all $0\leq i\leq n-1$. Hence, choosing $W=\lceil\sqrt{n}\rceil$ in the Spline Lemma (Lemma~\ref{lem:spline}), we have for all sufficiently large $n$ that $\fmarg, \phi_i\in\cN(\Theta(\sqrt{n)},\Theta(1))$. Next, observe that $g_1\in\cN(4,2)$. This follows from the fact that $g_1(x)=\sfm_2(\relu(\sfm_1(x)))$, where $\sfm_1(x)=[1 \; 1\;  1]^Tx+[0\; -\frac{1}{2}\; -1]^T$, and $\sfm_2(x)=[2\; -4\;\; 2]^Tx$. Hence, by recalling that $g_s=\underbrace{g_1\circ g_1\circ\ldots\circ g_1}_{s\text{ times}}$, the repeated application of Compose Lemma (Lemma~\ref{lem:compose}) yields $g_s\in\cN(3s+1,s+1)$. Therefore, an application of Compose Lemma (Lemma~\ref{lem:compose}) yields that for any $0\leq i\leq n-1$, $\phi_i(g_s(n\fmarg(\cdot)-i))\in\cN(3s+\Theta(\sqrt{n}),s+\Theta(1))$. Finally, by applying the Add Lemma (Lemma~\ref{lem:seradd}), we get $\sum_{i=0}^{n-1}\phi_i(g_s(n\fmarg(\cdot)-i))\in\cN(\Theta(n^{\frac{3}{2}})+s\Theta(n),ns+\Theta(n))$. 

Now, recalling that $\fmarg\in\cN(\Theta(\sqrt{n}),\Theta(1))$, a use of the Parallel Lemma (Lemma~\ref{lem:parallel}) gives $\phi\in\cN(\Theta(n^{\frac{3}{2}})+s\Theta(n),ns+\Theta(n))$. By Lemma~\ref{lem:map}, since a choice of $s=\max\{1,\lceil\log(\frac{2\sqrt{2}}{n\epsilon})\rceil\}$ yields $\mathsf{W}(\phi\# U,P)\leq \epsilon$, we have $N^*(n,\epsilon)\leq O(n^{\frac{3}{2}})+\max\{1,\lceil\log(\frac{2\sqrt{2}}{n\epsilon})\rceil\}O(n)$, which completes the proof of Theorem~\ref{th:main}.


\section{The Converse Part: Lower Bound on the number of neurons}

\label{sec:conv}

It is difficult to directly obtain lower bounds $N^*(n,\epsilon)$. Instead we focus on a related quantity, the minimum number of neurons required by a neural network of a fixed depth $L$ to $\epsilon$-approximate histogram distributions. For this section, we shall be considering $\cE[0,1]_n^d$ for any general $d\geq 2$, and all definitions from Section~\ref{sec:prelims} are modified accordingly. For instance, we now consider neural nets whose output dimension is $d$, \textit{i.e.}, $\phi:\real\to\real^d$. 

Formally, define $N^*(n,\epsilon,L)$ to be the minimum achievable size for a network of depth $L$ to $\epsilon$-approximate $\cE[0,1]_n^d$. The following theorem states the lower bound $N^*(n,\epsilon,L)$.

\begin{table*}[t!]
\centering
\begin{tabular}{c|c|c}
\multirow{2}{*}{Regime} & \multicolumn{2}{c}{Optimality}\\\cline{2-3}
                     & Our construction & Construction in \cite{PSB}\\
\hline
\hline
$\epsilon\to0$ & \multirow{2}{*}{Yes} & \multirow{2}{*}{Yes}\\
$n$ fixed &  & \\
\hline
$n\to\infty$ & \multirow{2}{*}{Not known} & \multirow{2}{*}{Not known}\\
$\epsilon$ fixed &  & \\
\hline 
$n\to\infty$ & \multirow{2}{*}{Not known} & \multirow{2}{*}{Not known}\\
$\epsilon=\Theta(n^{-\theta}), \theta>1$ &  & \\
\hline
$n\to\infty$ & \multirow{2}{*}{Yes} & \multirow{2}{*}{Not known}\\
$\epsilon=\Theta(2^{-n})$ &  & \\
\end{tabular}
\caption{Table documenting the optimality of different constructions for different regimes of $\epsilon$ and $N$. `Not known' means that the construction does not meet the bound in Theorem~\ref{th:conv}.}
\label{tab:conv}
\end{table*}

\begin{theorem}\label{th:conv}
Fix $n$, $\epsilon>0$, $L\geq 1$. The minimum achievable size for a network of depth $L$ to $\epsilon$-approximate $\cE[0,1]_n^d$ satisfies
$$
N^*(n,\epsilon,L)\geq\max\biggl\{L,L\biggl[\frac{1}{e}\biggl(\frac{C(d)}{n\epsilon}\biggr)^{\frac{d-1}{L}}-1\biggr]\biggr\},
$$
where $C(d)\triangleq \frac{d}{d-1}\biggl(\frac{2\Gamma(\frac{d+1}{2})\Gamma(\frac{3}{2})}{\pi^{\frac{d}{2}}\sqrt{d}}\biggr)^{\frac{1}{d-1}}$.
\end{theorem}

Based on Theorem~\ref{th:conv}, and setting $d=2$, Table~\ref{tab:conv} marks whether our neural network in Section~\ref{sec:const} and the network constructed in \cite{PSB} are optimal for various regimes of $\epsilon$ and $n$. Here, optimality implies whether the neural net $\epsilon$-approximating $\cE[0,1]_n^2$ using some depth $L$ has its size $N$ to be equal in orders of magnitude to the bound on $N^*(n,\epsilon,L)$ provided in Theorem~\ref{th:conv}. For example, we see that in the regime where $\epsilon\to 0$ and $n$ is fixed, both our construction and that of \cite{PSB} are optimal. However, in most regimes we are unable to guarantee optimality. We believe the problem stems from the weakness in lower bound. For example, consider the regime where $\epsilon$ is fixed and $n\to \infty$. Using a depth $L=\Theta(1)$, the bound in Theorem~\ref{th:conv} reduces $N^*(n,\epsilon,L)\geq \Theta(1)$. This is clearly loose, since a constant sized network should not be able to $\epsilon$-approximate $\cE[0,1]_n^2$ as $n$ grows. 

The proof of Theorem~\ref{th:conv} is an extension of a method used in \cite{BT}. A well-known property of a function $f:\real\to\real^d$ which is realized by a ReLU net is that the domain of $f$ can be partitioned into at most some $\zeta$ convex parts, such that $f$ is affine when restricted to each of these parts. In other words, the image of $f$ in $\real^d$ is supported in at most $\zeta$ $1$-dimensional affine hyperplanes in $\real^d$. The following lemma from \cite{BT} gives a bound on the number $\zeta$ based on the size $N$ and depth $L$ of the network.

\begin{lemma}\label{lem:numparts}
Let $f:\real\to\real^d$ is realized by a ReLU net of size $N$ and depth $L$. Then, the image of $f$ is supported on at most $\zeta$ $1$-dimensional affine hyperplanes on $\real^d$, where
$$
\zeta\leq (e(\frac{N}{L}+1))^L.
$$
\end{lemma}

\begin{IEEEproof}
See Lemma~2.1 of \cite{BT}. 
\end{IEEEproof}
The following theorem bounds the distance between $n$-tiled histogram distribution from $\cE[0,1]_n^d$ and any distribution supported on $\zeta$ $1$-dimensional affine hyperplanes in $\real^d$.

\begin{theorem}\label{th:lbhyp}
Let $P\in\cE[0,1]_n^d$ and $Q$ be a distribution supported on any $\zeta$ $1$-dimensional affine hyperplanes in $\real^d$. Then 
$$
\mathsf{W}(P,Q)\geq C(d)\zeta^{-\frac{1}{d-1}}n^{-1},
$$
where $C(d)\triangleq \frac{d}{d-1}\biggl(\frac{2\Gamma(\frac{d+1}{2})\Gamma(\frac{3}{2})}{\pi^{\frac{d}{2}}\sqrt{d}}\biggr)^{\frac{1}{d-1}}$.
\end{theorem}
It is easy to see that Theorem~\ref{th:conv} follows by simply plugging in the bound on $\zeta$ from Lemma~\ref{lem:numparts} to the expression in Theorem~\ref{th:lbhyp}, and noting the trivial bound $N\geq L$. The rest of this section is thus dedicated to proving Theorem~\ref{th:lbhyp}. 

We begin by introducing a key technical result from \cite{BT}. For any $B\subseteq\real^d$, denote by $\diam(B)$ and $\vol(B)$ respectively, the diameter\footnote{The diameter of a set is defined as $\diam(B)=\sup_{x,y\in B}||x-y||$.} and the volume of $B$. Given any set $\Lambda\subseteq\real^d$ and any distribution $Q$, define 
\begin{align*}
\mathsf{W}(Q,\Lambda) & =\inf_{\pi\in\Pi(Q,\Lambda)}\int_{\real^d\times\real^d}||x-y||d\pi(x,y)\\
                      &=\inf_{\gamma\in\Pi_\Lambda}\mathsf{W}(\gamma,Q),
\end{align*}
where $\Pi(Q,\Lambda)$ is the set of all couplings between $Q$ and distributions supported on $\Lambda$, and $\Pi_\Lambda$ is the set of all distributions supported on $\Lambda$. The following lemma from \cite{BT} bounds the distance $\mathsf{W}(U_B,S)$ between the uniform distribution on any bounded set $B$ in $\real^d$ and distributions supported on a $1$-dimensional hyperplane $S$.

\begin{lemma}\label{lem:telg}
Let $B\subseteq\real^d$ be a bounded set, and let $U_B$ be the uniform measure on $B$. Let $S$ be any $1$-dimensional hyperplane in $\real^d$. Then, we have
$$
\mathsf{W}(U_B,S)\geq \frac{d}{d-1}\biggl(\frac{2\Gamma(\frac{d+1}{2})\Gamma(\frac{3}{2})}{\pi^{\frac{d}{2}}\diam(B)}\vol(B)\biggr)^{\frac{1}{d-1}}.
$$
\end{lemma}
\begin{IEEEproof}
See Lemma A.3 of \cite{BT}.
\end{IEEEproof}

\begin{IEEEproof}[Proof of Theorem~\ref{th:lbhyp}]
Let $S_1, S_2, \cdots,S_\zeta$ be the $\zeta$ hyperplanes on which $Q$ is supported, and define $S=\cup_{i=1}^\zeta S_i$. Define the sets $c_\bk(j)\triangleq\{x\in c_\bk: j=\argmin_{1\leq l\leq\zeta}||x-S_l||\}$, $\bk\in\{0,1,\cdots,n-1\}^d, 1\leq j\leq\zeta$, where $||x-S_j||=\min_{y\in S_j}||x-y||$.\footnote{The minimum exists since hyperplanes are closed sets.} Therefore, 
\begin{align}
\mathsf{W}(P,Q) & \geq \inf_{\pi\in\Pi(P,S)}\int_{\real^d\times\real^d}||x-y||d\pi(x,y)\notag\\
                & = \inf_{\pi\in\Pi(P,S)}\int_{\cup_{\bk}c_\bk\times\real^d}||x-y||d\pi(x,y)\notag\\
                &=\inf_{\pi\in\Pi(P,S)}\sum_{\bk}\int_{c_\bk\times\real^d}||x-y||d\pi(x,y), \label{eq:lower:1}
\end{align}
where the last equality follows using the fact that $P$ is absolutely continuous with the Lebesgue measure on $\real^d$.\footnote{This follows from Definition~\ref{def:hist}.} Now, for every $x\in c_\bk$, let $S_j$ be the hyperplane with the least index $j$ that is closest to $x$. The fact that $S_j$ is closed and convex implies that there exists a unique $y\in S_j$ satisfying $||x-y_x||=\argmin_{y\in S_j}||x-y||$. Then for any coupling $\pi^*\in\Pi(P,S)$ that only puts weight in the set $\{(x,y_x):x\in\real^d\}$, we have 
\begin{align}
\inf_{\pi\in\Pi(P,S)} & \sum_{\bk}\int_{c_\bk\times\real^d}||x-y||d\pi(x,y)\notag\\
&=\inf_{\pi\in\Pi(P,S)}\sum_{\bk}\int_{c_\bk\times\real^d}||x-y_x||d\pi^*(x,y)\notag\\
&=\sum_{\bk}\sum_{j=1}^\zeta\int_{c_\bk(j)}||x-S_j||dP(x), \label{eq:lower:2}
\end{align}
where the last equality uses the absolute continuity of $P$ with repect to the Lebesgue measure on $\real^d$. Plugging \eqref{eq:lower:2} in \eqref{eq:lower:1}, we have 
\begin{align}
\mathsf{W}(P,Q)& \geq \sum_{\bk}\sum_{j=1}^\zeta\int_{c_\bk(j)}||x-S_j||dP(x)\notag\\
&=\sum_{\bk}\sum_{j=1}^\zeta w_\bk\int_{c_\bk(j)}||x-S_j||dx, \label{eq:lower:3}
\end{align}
where the last equality uses the Definition~\ref{def:hist}.

Next, note that $S_j$ is closed and convex, i.e., every $x\in\real^d$ has a unique point $y_x$ is $S_j$ which minimizes the distance. Let $B$ be any set and let $U_B$ be the uniform measure on it. Then, by choosing a coupling $\pi$ which is supported on $\{(x,y_x):x\in\real^d\}$, we have $\mathsf{W}(U_B,S_j)=\frac{1}{\vol(B)}\int_{B}||x-S_j||dx$. Plugging this in \eqref{eq:lower:3}, we have
\begin{align}
\mathsf{W}(P,Q) & \geq \sum_{\bk}\sum_{j=1}^\zeta w_\bk\vol(c_\bk(j))\mathsf{W}(U_{c_\bk},S_j)\notag\\
         & \geq \sum_{\bk}\sum_{j=1}^\zeta w_\bk\vol(c_\bk(j))\notag\\
         &\hspace{1cm}\frac{d}{d-1}\biggl(\frac{2\Gamma(\frac{d+1}{2})\Gamma(\frac{3}{2})}{\pi^{\frac{d}{2}}\diam(c_\bk(j))}\vol(c_\bk(j))\biggr)^{\frac{1}{d-1}}\label{eq:lower:4}\\
         & \geq \sum_{\bk}\sum_{j=1}^\zeta C(d)w_\bk\vol(c_\bk(j))(n\vol(c_\bk(j)))^{\frac{1}{d-1}}\label{eq:lower:5}\\
         & \geq \sum_{\bk}C(d)w_\bk n^{\frac{1}{d-1}}\zeta\biggl(\frac{1}{\zeta}\sum_{j=1}^\zeta\vol(c_\bk(j))\biggr)^{\frac{d}{d-1}}\label{eq:lower:6}\\
         & \geq \sum_{\bk}C(d)w_\bk n^{\frac{1}{d-1}}\zeta^{-\frac{1}{d-1}}\vol(c_\bk)^{\frac{d}{d-1}}\notag\\
         & \geq C(d)\zeta^{-\frac{1}{d-1}}\sum_{\bk}w_\bk n^{-(d+1)}\notag\\
         & \geq C(d)\zeta^{-\frac{1}{d-1}}n^{-1}\label{eq:lower:7},
\end{align}
where \eqref{eq:lower:4} follows using Lemma~\ref{lem:telg}, \eqref{eq:lower:5} uses the fact that $\diam(c_\bk(j))\leq\diam(c_\bk)=\frac{\sqrt{d}}{n}$, \eqref{eq:lower:6} follows using Jensen's inequality, \eqref{eq:lower:7} uses the fact $\sum_k w_k=n^d$ from Definition~\ref{def:hist}.
\end{IEEEproof}

\section{Concluding remarks}\label{sec:conc}

In this work, we studied the problem of approximating $n$-tiled histogram distributions by ReLU neural networks with uniform seed. We obtained a new upper bound on the achievable size of the network, which is tighter than existing results. The main ingredient we leveraged to tighten the bound is an efficient construction of neural networks representing free knot linear splines studied in \cite{DDFHP}. We also computed a lower bound on achievable size, but unfortunately the bounds do not match. We believe that the lower bound is loose and tightening it shall be a future direction of research. 

\bibliographystyle{IEEEtran}
\bibliography{ITW_2020}

\end{document}